\documentclass[a4paper, english, 11pt]{article}

\usepackage{geometry}
\usepackage{babel}

\usepackage{setspace}
\usepackage{parskip}

\usepackage{natbib}
\usepackage{graphicx}
\graphicspath{ {GFX/} }

\usepackage[inline]{enumitem}

\usepackage{booktabs}
\usepackage{siunitx}
\usepackage[flushleft]{threeparttable}
\usepackage{caption}
\usepackage{subcaption}
\usepackage{setspace}
\usepackage{rotating}
\usepackage{array}
\usepackage{adjustbox}
\usepackage{url}
\usepackage{makecell}
\usepackage{multirow}
\usepackage{tabularx}

\usepackage{mathrsfs}
\usepackage{amsmath}
\usepackage{amssymb}
\usepackage{amsbsy}
\usepackage{amsthm}
\usepackage{amsfonts}
\usepackage{mathabx}
\usepackage{bm}
\usepackage{dcolumn}
\usepackage{multirow} 
\usepackage{booktabs}
\usepackage{makecell}
\usepackage{siunitx}
\usepackage{scalerel}

\usepackage{calrsfs}
\usepackage{bbm}

\usepackage{xcolor}

\makeatletter
\newcommand{\xleftrightarrow}[2][]{\ext@arrow 3359\leftrightarrowfill@{#1}{#2}}
\newcommand{\xdashrightarrow}[2][]{\ext@arrow 0359\rightarrowfill@@{#1}{#2}}
\newcommand{\xdashleftarrow}[2][]{\ext@arrow 3095\leftarrowfill@@{#1}{#2}}
\newcommand{\xdashleftrightarrow}[2][]{\ext@arrow 3359\leftrightarrowfill@@{#1}{#2}}
\def\rightarrowfill@@{\arrowfill@@\relax\relbar\rightarrow}
\def\leftarrowfill@@{\arrowfill@@\leftarrow\relbar\relax}
\def\leftrightarrowfill@@{\arrowfill@@\leftarrow\relbar\rightarrow}
\def\arrowfill@@#1#2#3#4{%
$\m@th\thickmuskip0mu\medmuskip\thickmuskip\thinmuskip\thickmuskip
\relax#4#1
\xleaders\hbox{$#4#2$}\hfill
#3$%
}
\makeatother

\usepackage{mathtools}
\usepackage{algpseudocode, algorithm, algorithmicx}

\usepackage{setspace}
\let\Algorithm\algorithm
\renewcommand\algorithm[1][]{\Algorithm[#1]\setstretch{1.2}}

\algrenewcommand\algorithmicrequire{\textbf{Precondition:}}
\algrenewcommand\algorithmicensure{\textbf{Postcondition:}}
\algnewcommand{\LineComment}[1]{\State \(\triangleright\) #1}

\usepackage{titling}
\usepackage[condensed]{roboto}

\makeatletter

\renewcommand\section{\@startsection {section}{1}{\z@}%
{-3.5ex \@plus -1ex \@minus -.2ex}%
{2.3ex \@plus.2ex}%
{\normalfont\robotocondensed\Large}}

\renewcommand\subsection{\@startsection{subsection}{2}{\z@}%
{-3.25ex\@plus -1ex \@minus -.2ex}%
{1.5ex \@plus .2ex}%
{\normalfont\robotocondensed\large}}

\renewcommand\subsubsection{\@startsection{subsubsection}{3}{\z@}%
{-3.25ex\@plus -1ex \@minus -.2ex}%
{1.5ex \@plus .2ex}%
{\normalfont\robotocondensed\normalsize}}

\renewcommand\paragraph{\@startsection{paragraph}{4}{\z@}%
{3.25ex \@plus1ex \@minus.2ex}%
{-1em}%
{\normalfont\robotocondensed\normalsize}}

\renewcommand\subparagraph{\@startsection{subparagraph}{5}{\parindent}%
{3.25ex \@plus1ex \@minus .2ex}%
{-1em}%
{\normalfont\robotocondensed\normalsize}}

\makeatother

\usepackage{authblk}
 
\usepackage{xr}
\begin{document}

\title{Modeling non-linear Effects with Neural Networks in Relational Event Models}


\author[1]{Edoardo Filippi-Mazzola}
\author[1]{Ernst C. Wit}
\affil[1]{Institute of computing, Universit\`{a} della Svizzera italiana, Lugano, Swizterland}
\date{}


\renewcommand{\refname}{\large References}


\maketitle


\begin{abstract}
\noindent

Dynamic networks offer an insight of how relational systems evolve. However, modeling these networks efficiently remains a challenge, primarily due to computational constraints, especially as the number of observed events grows. This paper addresses this issue by introducing the Deep Relational Event Additive Model (DREAM) as a solution to the computational challenges presented by modeling non-linear effects in Relational Event Models (REMs). DREAM relies on Neural Additive Models to model non-linear effects, allowing each effect to be captured by an independent neural network. By strategically trading computational complexity for improved memory management and leveraging the computational capabilities of Graphic Processor Units (GPUs), DREAM efficiently captures complex non-linear relationships within data. This approach demonstrates the capability of DREAM in modeling dynamic networks and scaling to larger networks. Comparisons with traditional REM approaches showcase DREAM superior computational efficiency. The model potential is further demonstrated by an examination of the patent citation network, which contains nearly 8 million nodes and 100 million events.

\vspace{5mm}
\end{abstract}

\section{Introduction}
\label{sec:intro}

Dynamic network modeling have emerged as an essential tool in social network studies, providing a nuanced perspective on the evolving nature of interactions and relationships. These networks capture the dynamism inherent in dynamic interacting structures by shedding light on how connections form, dissolve, or transform over time. Although the inclusion of the temporal dimension increases the complexity of the models, it provides richer insights, revealing patterns that static networks might miss.

Relational Event Models (REMs) \citep{butts_2008,perry_point_2013,bianchi_review_2023} are an efficient and flexible framework for modeling dynamic networks, particularly in settings where events, or interactions between actors, occur sequentially over time. Unlike traditional network models, which focus on aggregated states or snapshots, REM focuses on micro-dynamics, tracking the chronological order of ties as they form or dissolve \citep{network_survey_kauerman_2020}. The great versatility of REMs is underscored by the diverse fields they have been applied to, including finance \citep{zappa_finance_2021,bianchi_finance_2021}, healthcare \citep{vu_relational_2017,amati_healthcare_2019}, and ecology \citep{patison_time_2015,tranmer_animal_2015,juozaitiene_ecological_2023}. Despite their easy adaptability, REMs practical utility is limited by their computational complexities \citep{welles_comunication_2014}. The computational difficulties associated with either the full or partial likelihood approach are acknowledged in the field, prompting researchers to seek alternative solutions. 

\cite{vu_relational_2015} first tackled the problem by proposing various sampling strategies on the risk-set connected to the partial-likelihood denominator. \cite{lerner_reliability_2020} demonstrated the robustness of REMs when the risk-set is sub-sampled via a nested-case-control approach, demonstrating that when REMs are used to model large dynamic networks, only one control per case is sufficient to obtain reliable estimates. This sub-sampling strategy was used by \cite{filippi-mazzola_stochastic_2023} to approximate the REM partial likelihood by a logistic regression, which reduces computational complexity and allows for efficient modeling of non-linear effects.

Non-linear approaches to REMs where first tackled by \cite{bauer_smooth_2022} using B-splines \citep{DeBoor1972}. By design, these spline-based models require to storing multiple high-dimensional model matrices. When these factors are combined with large networks with millions of dyadic interactions, many REM computing frameworks suffer from both convergence and memory management issues.

Inspired by Neural Additive Models \citep{agarwal_nam_2021}, we propose to model non-linear effects throguh a Deep Relational Event Additive Model (DREAM). DREAM strategically trades computational complexity with memory management by letting each effect be modelled by an independent neural network. By leveraging the higher computational power of Graphic Processor Units (GPUs), DREAM is able to capture complex non-linear effects among variables.  Each of these independent neural networks is trained at the same time using a Stochastic Gradient Descent (SGD) approach. SGD is especially renowned for its ability to handle large datasets and high-dimensional spaces as iteratively refines model parameters to ensure optimal convergence. The simultaneous estimation of these neural networks not only increases computational efficiency, but also ensures that interdependencies and mutual information among different effects are captured effectively. 

In this paper, we start by describing the methodological background on which REMs are built on in section~\ref{sec:back}. After defining how DREAM is structured in section~\ref{sec:DREAM}, we provide a comprehensive simulation study to test its robustness and efficiency in section~\ref{sec:sim_study}. To conclude, we show an application on the analysis of the US patent citation network in section~\ref{sec:patents_appl}.

\section{Background and Methods}
\label{sec:back}

REMs are a class of statistical models for sequences of social interaction events occurring over time. The primary focus of these models for is to model the pattern and structure of relationships that emerge as a series of observed social interactions or events. 

\subsection{Relational Event Model}
\label{subsec:rem}

In REMs, the primary statistical units are a series of recorded dyadic interaction defined as events. These are denoted as $e_i$ $(i=1,\dots,n)$ and are typically represented by the triplet $e_i=(s_i,r_i,t_i)$, which denotes that an action was initiated by a sender $s_i$, targeted towards a receiver $r_i$, and occurred at a specific time $t_i$. 

Following \cite{perry_point_2013}, it is possible to define a multivariate counting process $N_{sr}(t)$ that records the number of directed interactions between $s$ and $r$, up to time $t$, 
\begin{equation*}
	 N_{sr}(t) = \sum_{i\geq 1} \mathbf{1}_{\left\{t_i\leq t;~ s_i=s;~ r_i=r\right\}}.
\end{equation*}
$N_{sr}(t)$ is then a local submartingale, where, through Doob-Meyer decomposition $N_{sr}(t) = \Lambda_{sr}(t)+M_{sr}(t)$, we can define its predictable increasing process $\Lambda_{sr}(t)$. REMs describe this predictable increasing process by assuming that hat the counting process is an inhomogeneous Poisson proces, i.e., 
\begin{equation*}
	\Lambda_{sr}(t) = \int_{0}^{t}\lambda_{sr}(\tau) d\tau,
\end{equation*}
where $\lambda_{sr}$ is the hazard function of the relational event $(s, r)$. Considering the history of prior events $\mathcal{H}_{t}$ up to time $t$, a common method for modeling this intensity function relies on the log-linear model \citep{cox_regression_1972}. Consequently, the intensity function is expressed as the product of a baseline hazard $\lambda_0(t)$ and an exponential function of $q$ $\mathcal{H}_{t}$-measurable covariates $x$,
\begin{equation}
	\label{eq:rem}
	\lambda_{sr}(t \mid \mathcal{H}_{t}) = \lambda_0(t)e^{\sum_{k=1}^{q} f_k(x_k)},
\end{equation}
where $f_k(x_k)$ in the original formulation is modelled as a linear function weighted by a coefficient $\beta_k$, such that $f_k(x_k)= x_k \beta_k$.  

Given the network prior history, the model definition assumes conditional independence of events. Incorporating covariates into this structural design allows for an in-depth investigation into a multitude of individual influences or factors that contribute to the occurrence of the event. These influential components are typically classified as exogenous or endogenous. Exogenous factors generally pertain to attributes or effects related to the sender or receiver, offering insights into external dynamics. These could include individual characteristics, roles within the network, or external circumstances that influence their actions. On the other hand, endogenous factors relate to the intrinsic micro-mechanisms within the network itself. These are patterns or tendencies that arise from the inherent structure and dynamics of the network, becoming visible as the series of events progressively unfold over time. By recognizing and studying these factors, we can gain an understanding of how the network internal mechanisms shape the unfolding of events.

A further characteristic that lends significant appeal to the proportional hazard model is its absence of distributional assumptions concerning activity rates. This flexibility is a notable advantage over fully parametric models and enables to treat the baseline hazard $\lambda_0$ as a nuisance parameter \citep{cox_partial_1975}. This strategic consideration helps simplify the computational complexities that emerge when trying to estimate the weights in the full-likelihood derived from~\eqref{eq:rem}. The resulting partial likelihood is expresses as follows,
\begin{equation}
	\label{eq:pl}
	L_P(\beta) = \prod_{i=1}^{n} \left(\frac{\exp\left\{ \sum_{k=1}^{q} f_k(x_{s_ir_ik})  \right\}}{\sum_{(s_i^*,r_i^*)\in \mathcal{R}(t_i)}\exp\left\{\sum_{k=1}^{q} f_k(x_{s_i^*r_i^*k}) \right\}}\right),
\end{equation}
$\mathcal{R}(t)$ is defined as the risk-set, i.e., the set of all possible events that could have happend until time $t$. 

\subsection{Nested case control sampling}
\label{subsec:nested_case_control}

While the application of partial likelihood in (\ref{eq:pl}) introduces significant simplifications to REM estimation, its practical application is constrained by the dimensionality of its denominator. The risk set $\mathcal{R}(t)$ tends to increase quadratically with the number of nodes in traditional longitudinal networks, altough it may vary depending on the specific context. For instance, in citation networks, the risk set tends to expand linearly as new nodes cite existing documents within the network \citep{vu_dynamic_2011,filippi-mazzola_stochastic_2023}. However, irrespective of the individual scenarios under analysis, scalability remains a limiting factor for these models. Large networks, comprising millions of nodes, will inevitably pose computational challenges and potentially limit the model efficacy in such contexts.

A solution to this issue has been proposed by \cite{vu_relational_2015}, who introduced the idea of nested case-control sub-sampling the risk set \citep{borgan_methods_1995}. The central idea revolves around analyzing all the observed events, or ``cases," while only scrutinizing a smaller subset of non-events, termed ``controls." \cite{borgan_methods_1995} demonstrated that using a nested case-control sampled risk set yields a consistent estimator. Building on this concept, \cite{boschi_smooth_2023} and \cite{filippi-mazzola_stochastic_2023} extended the empirical findings of \cite{lerner_reliability_2020} to argue that in scenarios with a large number of nodes, one control per case is sufficient to obtain reliable parameter estimates.

When only a single control per case is considered, the partial likelihood in (\ref{eq:pl}) results in the likelihood of a logistic regression model where only successful outcomes are observed as responses. This key insight further enhances the practicality of REMs, reducing the computational complexity of estimating large and complex risk sets. With this transformation in place, the sub-sampled case-control version of the partial likelihood in (\ref{eq:pl}) is given as,  
\begin{equation}
	\label{eq:samp_pl}
	\widetilde{L_P}(\beta) = \prod_{i=1}^{n} \left[ \frac{\exp\left\{  f(x_{s_ir_i}) -  f(x_{s^*_ir^*_i}) \right\}}{1+\exp\left\{ f(x_{s_ir_i}) -  f(x_{s^*_ir^*_i}) \right\}}\right],
\end{equation}
where $f(x_{s_ir_i})= \sum_{k=1}^{q} f_k(x_{s_ir_ik})$, and $x_{s^*_ir^*_i}$ is a non-event for $i$th event with ranomly sampled sender $s_i^*$ and sampled receiver $r_i^*$ from $\mathcal{R}(t_i)$.

\section{Deep Relational Event Additive Model}
\label{sec:DREAM}

Altough standard REM formulations assume that the rate of interaction between $s$ and $r$ exhibits a linear dependence on the covariates, the relationship between predictors and event rates could be non-linear and exhibit a greater degree of complexity. If this is the case, deploying linear effects could inadvertently result in model oversimplification and the production of biased estimates. Consequently, these potential inaccuracies highlight the necessity of exploring alternative modeling techniques beyond the traditional linear framework. In this section, we propose the Deep Relational Event Additive Model (DREAM) to estimate non-linear effects in large networks, that leverages machine-learning methods and graphical processors units for efficient computation.

\subsection{Non-linear modeling with Neural Networks}
\label{subsec:non-linear}

Modeling non-linear effects in REMs was first tackled by \cite{bauer_smooth_2022} and \cite{filippi-mazzola_stochastic_2023}. Both heavily rely on the use of B-splines \citep{DeBoor1972}, a computational technique that represents data as a series of interconnected piecewise polynomial functions. While splines add value in accurately capturing non-linear patterns, their implementation comes with challenges, especially when it comes to the memory demands associated with the fitting procedure as each additional effect necessitates the creation of a multi-dimensional matrix.

\begin{figure}[h]
	\centering
	\includegraphics[width=\linewidth]{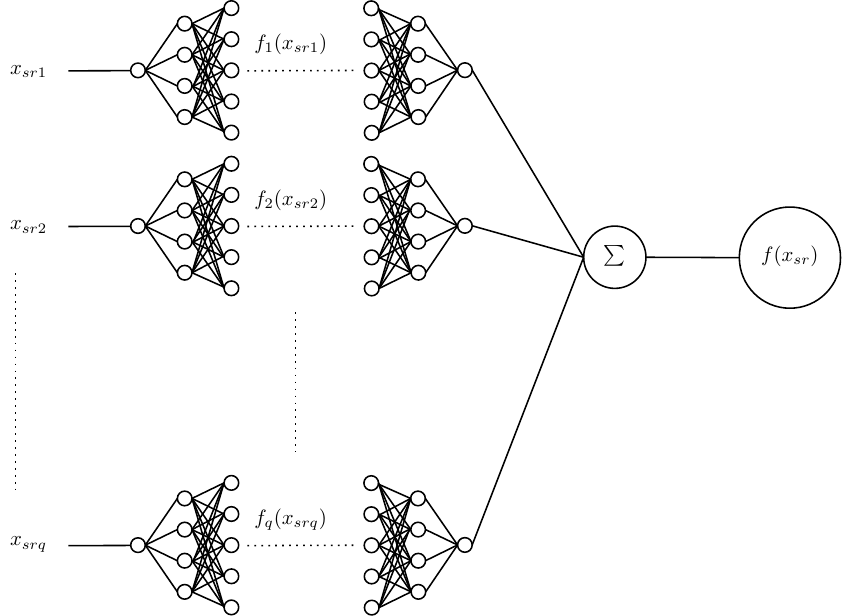}
	\caption{DREAM framework. Each effect is modeled via an independent ANN structure that captures its non-linearity.}
	\label{fig:DREAM_framework}
\end{figure}

DREAM strategically trades off memory usage with computational complexity. Following the recent developments of Neural Additive Models \citep{agarwal_nam_2021}, DREAM leverages multi-layered neural networks to model non-linear effects, where each effect is modeled by an independent neural network. Let then $f_k$ be a feed-forward Artificial Neural Network (ANN) \citep{ripley_pattern_1996} with a single input and a single output, separated by $L$ layers, for $l=1,\dots,L$. Each of these layers contains $m_1,\dots,m_L$ neurons. The output of each $f_k$ is the result of a series of sequential operations, such as 
\begin{align*}
	a_k^{(1)} &= \phi\left(\beta_k^{(1)}x_{srk} + \beta_{0k}^{(1)}\right)\\
	a_k^{(2)} &= \phi\left(\beta_k^{(2)}a_k^{(1)} + \beta_{0k}^{(2)}\right)\\
	&\vdots \\
	a_k^{(L-1)} &= \phi\left(\beta_k^{(L-1)}a_k^{(L-2)} + \beta_{0k}^{(L-1)}\right)\\
	a_k^{(L)} &= \beta_k^{(L)}a_k^{(L-1)} + \beta_{0k}^{(L)},
\end{align*}
where $a_k^{(l)}$ represents the output of the $l$-th layer for the $k$-th covariate, $\beta_k^{(l)}$ is the weight matrix and $\beta_{0k}^{(l)}$ is the bias. $\phi$ is a non-linear function, commonly referred as activation function. $f_k$ can then be expressed as 
\begin{equation}
	\label{eq:ANN}
	f_k(x_{srk}) =\beta_k^{(L)}\left( \ldots  \phi\left(\beta_k^{(2)}\left( \phi\left( \beta_k^{(1)}x_{srk} + \beta_{0k}^{(1)} \right) \right) + \beta_{0k}^{(2)} \right) \ldots \right)+\beta_{0k}^{(L)}.
\end{equation}
Let then $f(x_{sr})$ represent the collective sum of $q$ independent ANN output effects for $x_{srk}(t)$, with $k=1,\dots,q$ --- i.e., $f(x_{sr})=\sum_{k=1}^q f_k(x_{srk})$. Each of these ANNs is then trained simultaneously to maximize~\eqref{eq:samp_pl}. Fig.~\ref{fig:DREAM_framework} describes the structure of how the information is passing through $f(x_{sr})$, offering a clear understanding of the DREAM framework.
The most significant asset of this modeling technique lies in its interpretability. Through a visual examination of the individual shape functions, one can develop a comprehensive understanding of the model behavior when processing each effect, thereby mimicking the interpretability level of splines. In this regard, DREAM benefits in the identification of individual contributions and implications of each effect. Although this technique increases the computational complexity for evaluating the likelihood in~\eqref{eq:samp_pl} as the passage from one layer of the network to another requires multiple matrix multiplications, it eliminates the heavy memory usage associated with basis transformations. While modern frameworks allows efficient matrix operations, the efficiency of this approach is mainly guaranteed by recent technological advancements in the application of vectorized computations on Graphic Processor Units (GPUs). 

It is possible to extend this modeling technique with interaction effects simply by letting $f_k$ have a two-dimensional input, such as $f_k(x_{srk_1},x_{srk_2}))$, and an output of one dimension. Furthermore, we note that $f(x_{sr})$ can be interpreted as a proxy to an Additive Model  \citep{hastie_gams_1986}, where each $f_k$ is estimated by a smooth therm.


\subsection{Oscillatory activation functions}
\label{subsec:acivation_function}

ANN models use specific non-linear functions, known as activation functions, that facilitate the transformation of each neuron input into a non-linear output. Among the prevalent choices for these functions lies the family of Rectified Linear Units (ReLU) \citep{agarap_relu_2019}, known for their simplicity and computational efficiency. Despite the attractive features, ReLU functions are often affected by the issue known as the ``dying ReLU" \citep{Lu_dying_2020}. This emerges in many empirical applications when certain neurons within the network become perpetually inactive, i,.e., they continuously output zeros for specific regions of the input support space. This behavior makes the affected neurons essentially irrelevant during the training phase as once a neuron enters in this state, the gradient at that point becomes zero. Consequently, during the backpropagation phase, no updates are made to the weights connected to that neuron. The absence of any weight adjustment leads to a state of inertia where the neuron remains inactive, never contributing to the model again.

Within the family of non-linear activation functions, two noteworthy alternatives to the ReLU are the Sigmoid \citep{Narayan_sigmoiod_1997} and Hyperbolic Tangent (tanh) \citep{Namin_tanh_2009}. While both functions share similar sigmoidal curves, they exhibit distinct behaviors concerning their derivatives. Specifically, the sigmoid function derivative quickly approaches zero on both right and left sides. This behavior translates to smaller gradients, which in turn leads to protracted training periods. Furthermore, this rapid decline in the derivative magnitude opens to the vanishing gradient problem during backpropagation, which poses a significant challenge in achieving swift and stable convergence in the neural network.

In contrast, the tanh function mitigates some of these difficulties. Its derivative is characteristically sharper and maintains non-zero values over a more extended range on both ends. This design helps in alleviating the vanishing gradient problem to some extent. However, tanh is not without its limitations. Its adaptability is limited by its rigidity in defining the non-linear transformation shape. As a result, the tanh function sometimes struggles to model more intricate and nuanced patterns present in complex datasets.

The challenges associated with previously discussed activation functions prompted a rigorous exploration of alternative units. This led to the adoption of the Growing Cosine Unit (GCU) \citep{noel_gcu_2023}. Initially conceptualized to mitigate the ``dying ReLU" problem in convolutional neural networks, GCUs have emerged as a promising contender among oscillatory activation functions. Defined as
\begin{equation}
	\label{eq:gcos}
	\phi \left(a_k^{(l-1)}\right) = \left(\beta_k^{(l)}a_k^{(l-1)}+\beta_{0k}^{(l)}\right) \cos \left(\beta_k^{(l)}a_k^{(l-1)} + \beta_{0k}^{(l)}\right),
\end{equation}
where $a_k^{(l-1)}$ represents the output from the previous layer, GCUs exhibit a unique property. Unlike ReLU units, which typically yield a singular decision boundary, GCU neurons decision boundary comprises infinitely many parallel hyperplanes. This is attributable to the GCU activation function infinite zeros. Additionally, GCUs offer consistent and favorable derivatives, acting as a countermeasure against the vanishing gradient issue. Furthemore, this trait produces a more efficient training process, marked by reduced duration and improved convergence rates.

\subsection{Estimation}
\label{subsec:estimation}

Like most common machine-learning techniques, DREAM scalability in the estimation process is attained thanks to the adoption of a Stochastic Gradiend Descent (SGD) approach. SGD is particularly effective for large datasets and complex models, as it updates model parameters iteratively based on a subsets (batches) of data, rather than the entire dataset.

Within the different SGD methodologies, we specifically utilize the ADAM optimizer \citep{kingma_adam_2017}. ADAM stands out from other SGD approaches for its proven scalability and convergence reliability \citep{Reddi_adamconv_2018}. It calculates individual adaptive learning rates based on estimates of the first and second moments of the gradients. Consider $\nabla \widetilde{L_P}(\beta)_b$ as the gradient of the partial likelihood for batch $b$. In the ADAM optimization process, the first and second moment estimations are updated as follows:
\begin{align*}
	m_b &\gets \xi_1 m_{b-1} + (1-\xi_1)\nabla \widetilde{L_P}(\theta)_b, \\
	v_b &\gets \xi_2 v_{b-1} + (1-\xi_2)\nabla \widetilde{L_P}(\theta)_b^2,
\end{align*}
where $m$ and $v$ represent the first and second moment gradients, respectively. The hyperparameters $\xi_1$ and $\xi_2$ are instrumental in determining the extent to which past gradients influence the current moment updates.

ADAM incorporates bias correction to adjust for the initial bias in the first and second moments of the gradients. This correction is crucial because the moving averages of these gradients start from zero, leading to an initial bias toward zero, particularly noticeable at the early stages of training. To counteract this, ADAM modifies the moving averages with a correction factor that is directly related to the learning rate and inversely related to the iteration count. Denoting the current training step as $s$, the first and second moments undergo bias correction as follows:
\begin{align*}
	\hat{m}_{b,s} & = \frac{m_b}{1-\xi_1^s}, \\
	\hat{v}_{b,s} & = \frac{v_b}{1-\xi_2^s},
\end{align*}
with $\xi_1^s$ and $\xi_2^s$ approaching zero as $s$ increases. Consequently, the model parameters are updated by:
\begin{equation*}
	\beta_b \leftarrow \beta_{b-1} - \psi \frac{\hat{m}_{b,s}}{\sqrt{\hat{v}_{b,s}}+\epsilon},
\end{equation*}
where $\psi$ denotes the learning rate, determining the step size of each parameter update, and $\epsilon$ is a small constant (typically $1e-8$) to avoid any division by zero. It is important to note that adjusting the learning rate during training can further refine the estimation of the weights. However, due to the complexity involved in determining an optimal learning rate decay, we chose to maintain a constant learning rate, denoted as $\psi$, throughout the training process in our simulation studies and application.

\subsection{Uncertainty estimation with Gaussian Process Regression}
\label{subsec:gpr}

Traditional methods of uncertainty estimation, well-suited to conventional statistical models, often fall short when applied to complex ANN models. A common solution for non-parametric models is to employ non-parametric Bootstrap \citep{efron_boot_1979}.  Such procedure estimates pointwise uncertainty intervals by generating a sufficient number of sampled copies from original dataset. For each repetition, the model is re-trained on a new ``bootstrapped" version of the original dataset. Consequently, each retrain yields distinct estimates of the non-linear effects. Particularly volatile areas will produce curves with diverging shapes. Once a sufficient number of repetitions are completed, confidence bands can be directly computed from the estimated curves using percentiles such as the 2.5th and 97.5th for a 95\% confidence interval. By averaging the different curves, it is possible to provide an estimate of the observed effect.

While bootstrapping offers great adaptability and flexibility, it is undeniably computationally demanding, particularly when considering the training requirements of neural networks. To address this, we propose to limit the number of bootstrap refits on which DREAM is trained and infer the confidence bands that would typically be derived through hundreds of bootstrap refits with a Gaussian Processes Regression (GPR) \citep{rasmussen_gaussian_2006} approach.

We assume that $\hat{f}_k \sim \mathcal{GP}(m_k,K_k)$ where $m_k$ represents the constant mean function of the Gaussian Process, while $K_k$ is a Radial Basis Function (RBF) kernel with the form 
\begin{equation}\label{eq:kernel}
	K_k = \exp \left(- \frac{\mid\mid x - x' \mid \mid}{2l^2} \right),
\end{equation}
where $l$ is a scale parameter, while $x'$ is a subsequent value of $x$. Computing the posterior mean and covariance matrix, we obtain an estimated mean function whose uncertainty is represented by the standard deviations computed by square rooting the diagonal of the posterior covariance matrix. The $O(n^3)$ computational complexity of GPR models presents a limitation for large-scale applications. This higher computational cost is due to the inversion of the kernel matrix in the posterior. However, to obtain pointwise estimates of the curves, it is sufficient to evaluate the kernel matrix on a reduced sample of equidistant points on the support range of $x$. Knowing the upper and the lower bound of each covariate, we can generate a vector $\Tilde{x}$ of equidistant points. We can then define $\Tilde{x}  = (\Tilde{x}_{1},\Tilde{x}_{2}\dots,\Tilde{x}_{N})$, where $\Tilde{x}_{i}  = x_{\min}+(i-1)\Delta$ for $i=1,\dots,N$, with $\Delta=\frac{x_{\max}-x_{\min}}{N-1}$ and $N<<n$. With $\Tilde{x}$, we can compute the posterior estiamtes of the Gaussian Process, 
\begin{align}
	\hat{\mu}(\Tilde{x})&=K_k^\top[K_k+I]^{-1}\hat{f}(\Tilde{x}),\\
	\hat{\Sigma}(\Tilde{x})&=K_k-K_k^\top[K_k+I]^{-1}K_k,
\end{align}
where $I$ is the identity matrix that to improve numerical stability during inversion. While in many scenarios it is possible to scale this identity matrix by multiplying $I$ to a constant, we have noticed in our applications that the results remain invariant to such scaling. Overall, this alternative approach is designed to offer a more computationally efficient alternative to a Bootstrap approach. 

\section{Simulation study}
\label{sec:sim_study}

In this section, we present a series of simulation studies that highlight DREAM ability in accurately identifying non-linear effects in dynamic networks. After having generated relational event data, the initial study delves into DREAM ability in reconstruct the true genearting functions behind observed effects.

Given their close similarity, it is natural to compare additive Neural Network models with basis-function-based additive models such as Generalized Additive Models (GAMs). Yet GAMs estimation encounters difficulties when faced with high-dimensional data. The following simulation study highlights DREAM effectiveness in accurately representing the true generating function and its high scalability, as evidenced by a direct comparison with the \texttt{gam} function from the MGCV package in R.

The full code has been written in python within the pytorch suite \citep{pytorch} and it is publicly available in a GitHub repository (\url{https://github.com/efm95/DREAM}) together with all the simulations and application.

\subsection{True functions recovery}
\label{subsec:func_rec}

\begin{figure}[h]
	\centering
	\includegraphics[width=\textwidth]{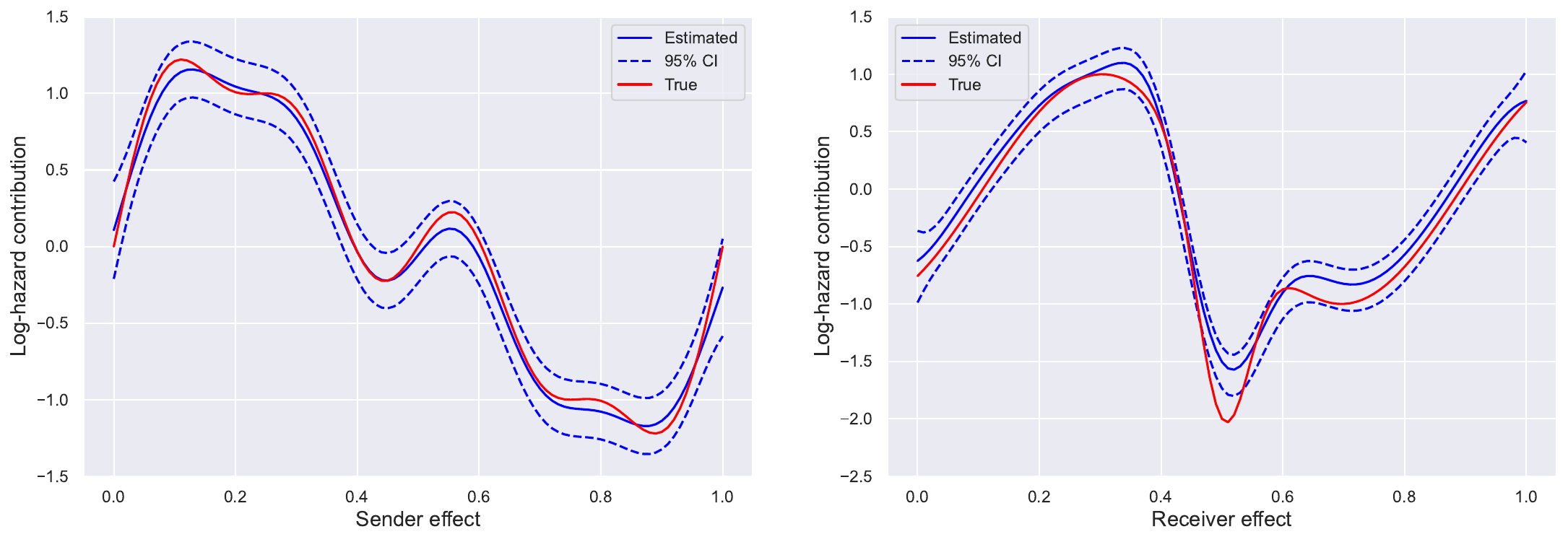}
	\caption{True and estimated effects along with their confidence intervals. The red lines denote the actual effects, whereas the blue lines are the estimated effects, and the dashed-blue lines represent the confidence intervals. Confidence intervals are calculated by adding and subtracting twice the standard deviation from the estimated mean functions.}
	\label{fig:sim_study_1}
\end{figure}

We simulated relational event data under the assumption that each node possesses both a sender and a receiver covariate ($\mathcal{U}(0,1)$). The effect of each covariate is given in red in Fig.~\ref{fig:sim_study_1}.

A network with 5,000 nodes and 500,000 edges is sampled. The fitted curve was estimated via five boostrap refits, followed by the application of a GPR model using the scikit-learn Python library \citep{scikit-learn}. The ANN architecture was determined via CV (details can be found in the supplementary materials in S1).

The results show how the estimated curves follow the true generating functions behavior, while the estimated confidence intervals obtained through this approach encompass the true functions over the variables support.

\subsection{Accuracy comparision with GAM}
\label{subsec:gam_comparison}

As previously introduced, similar non-linear effects can be modelled by using a logistic regression additive model approach. A good choice is the \texttt{gam} function within the MGCV package in R, where the smooth terms are estimated via penalized b-splines. The great advantage of MGCV is that the degrees of freedom of the splines are authomatically selected, thus reducing the number of hyperparameters that are required to be set. 

Comparing models using traditional information criteria such as AIC or BIC may not provide a fair assessment in the context of ANNs. This is because ANNs incorporate a substantially larger number of parameters compared to GAMs or other classical statistical models. To address this challenge, we adopt an alternative metric for model comparison. Specifically, in Table~\ref{tab:neg_PL}, we report the maximized log-partial-likelihood values obtained by DREAM and a GAM, alongside the log-partial-likelihood computed from the sampled population with the true generating functions. By using the Kullback–Leibler (KL) divergence assessed with the sampled populaiton, we provide a comparison that considers model fit and the ability to accurately reconstruct the true generating functions.

\begin{table}
	\centering
	\begin{tabular}{@{}cccc@{}}
		\toprule
					 &GAM  				& DREAM 	  & Population \\ \midrule
		$\log L_P$	& -232’991.59 & -232'102.28  		& -231'634.90 \\ 
		$KL(\text{Pop.} || \text{Model})$ & 1794.71		 &   1647.98	&  - \\ \bottomrule
	\end{tabular}
	\caption{Log-partial-likelihood values for each estiamted compared with the one computed from the sampled population using the true generating model. Subsequently, the KL-divergence values are assessed in relation to this same sampled population. }
	\label{tab:neg_PL}
\end{table}

From the results in Table~\ref{tab:neg_PL}, it is possible to notice DREAM attains a log-partial-likelihood score that more closely attains the one of the sampled population. As a consequence,  this is also reflected in its KL-divergence scores. For this scenario, GAM with smooth terms is slightly outperformed. However, the proximity of the performances between DREAM and GAM suggests that both models offer a similar level of accuracy in approximating the true model.



\subsection{Time efficiency}
\label{subsec:conv}

Estimations in the MGCV package use highly optimized Newtonian solvers, written in C routines to achieve rapid convergence. However, the computational efficiency of these solvers is frequently undermined by R less-than-optimal memory management system \citep{kotthaus_runtime_2015}. This results in computational bottlenecks, prolonging the time required for the algorithm to converge. In some instances, the inefficiency in memory management can even lead to computational overflow, further complicating the estimation process.

In contrast, DREAM relies on the PyTorch suite \citep{pytorch}, a deep learning framework that excels in handling vectorized operations. PyTorch is specifically designed to leverage the computational capabilities of Graphics Processing Units (GPUs). Using Google Colab free GPUs (Nvidia Tesla T4 with 15GB of memory), we run two sets of simulation to compare MGCV convergence times with DREAM with the implementation of the GPR approach to estimate the curves.

\begin{figure}
	\centering
	\begin{subfigure}{.5\textwidth}
		\centering
		\includegraphics[width=\linewidth]{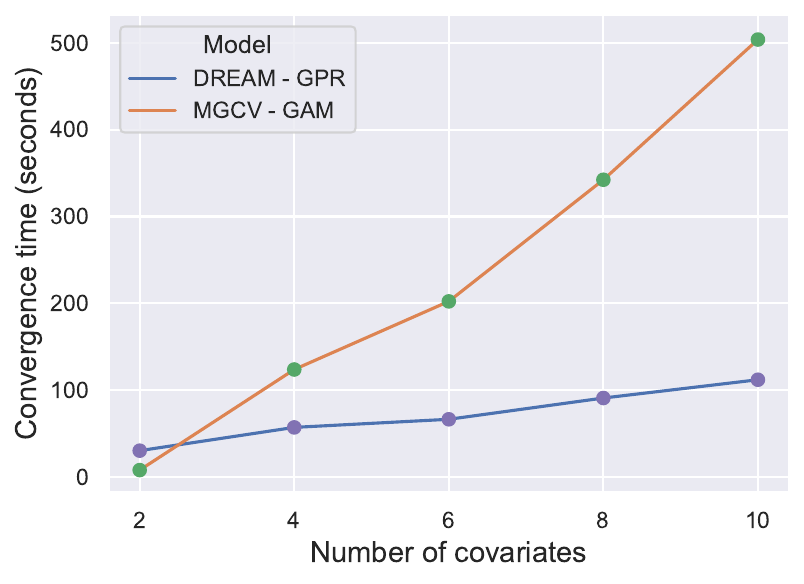}
		\caption{100'000 events with 1'000 actors.}
		\label{fig:sub1}
	\end{subfigure}%
	\begin{subfigure}{.5\textwidth}
		\centering
		\includegraphics[width=\linewidth]{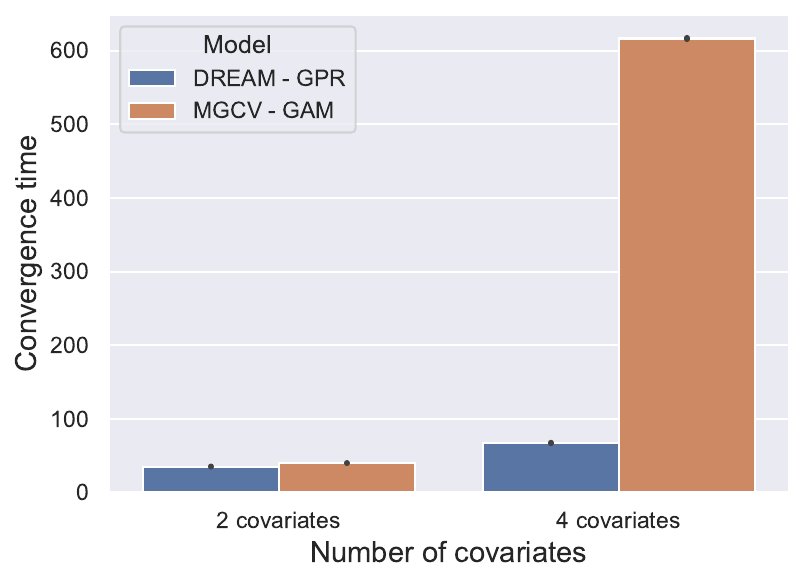}
		\caption{500'000 events with 5'000 actors.}
		\label{fig:sub2}
	\end{subfigure}
	\caption{Comparison of convergence times between MGCV and DREAM across two simulated REM data scenarios.}
	\label{fig:sim_study_2}
\end{figure}

Figure~\ref{fig:sub1} compares the convergence times of MGCV and DREAM using generated REM data that comprised 1'000 nodes and 100'000 events. We gradually augmented the complexity by sequentially covariates, thus increasing the number of non-linear effects that each model needed to estimate. We carried out the fitting procedure ten times. While MGCV convergence time initially appeared faster with only two covariates, its performance rapidly degraded as the complexity grows, revealing the computational bottlneck within R. Conversely, DREAM exhibited only a modest uptick in convergence time as the complexity increased.

Figure~\ref{fig:sub2} presents for a larger dataset comprising 5'000 nodes and 500'000 events. While the C routines lend MGCV stability in its convergence times, it becomes noticeably strained with the inclusion of 4 covariates, taking considerably longer. It is to note that with this size of data, we were not able to fit a model with more covariates as the algorithm failed to converge. 

\section{US patent citation network}
\label{sec:patents_appl}

To demonstrate DREAM practical applicability on large networks, we model non-linear effects in the US patent citation network that contains nearly 100 million citations and almost 8 million patents from 1976 to 2022. We chose this specific application not only because of its size and complexity, but also because the data preprocessing procedures and the computation of the statistics are well-defined, making the study more accessible and simple to replicate. The preprocessing of the patent citation network can be found at \url{https://github.com/efm95/STREAM}. A detailed model selection for the application to the US patent citation network is extensively covered in the supplementary material, under section S2.

In order for a patent to be formally issued, the applicant must disclose all relevant prior art. As a result, the US patent citation network consists of patents that cite earlier works in relation to their issuance date. This results in a dynamic network that is constantly growing and expanding. Within this network, nodes are represented by patents, and as they are published, they establish connections to pre-existing nodes in the network via citations. The aim of this mdoeling exercise is to identify what drives a patent s to cite a patent r at time t.

\begin{figure}
	\centering
	\begin{subfigure}{.33\textwidth}
		\centering
		\includegraphics[width=\linewidth]{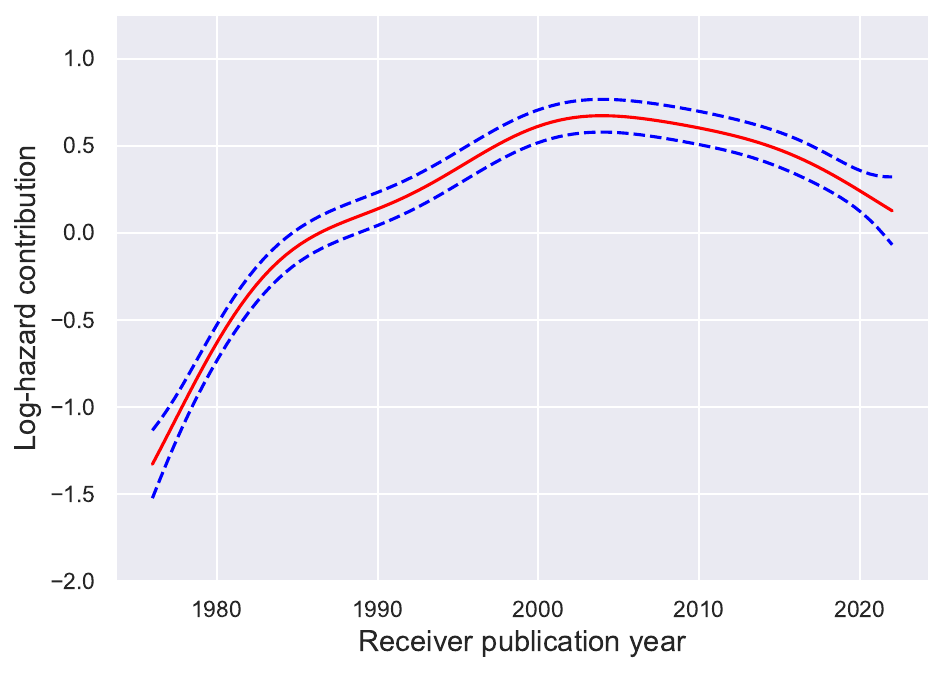}
		\caption{}
		\label{fig:pub_year}
	\end{subfigure}%
	\begin{subfigure}{.33\textwidth}
		\centering
		\includegraphics[width=\linewidth]{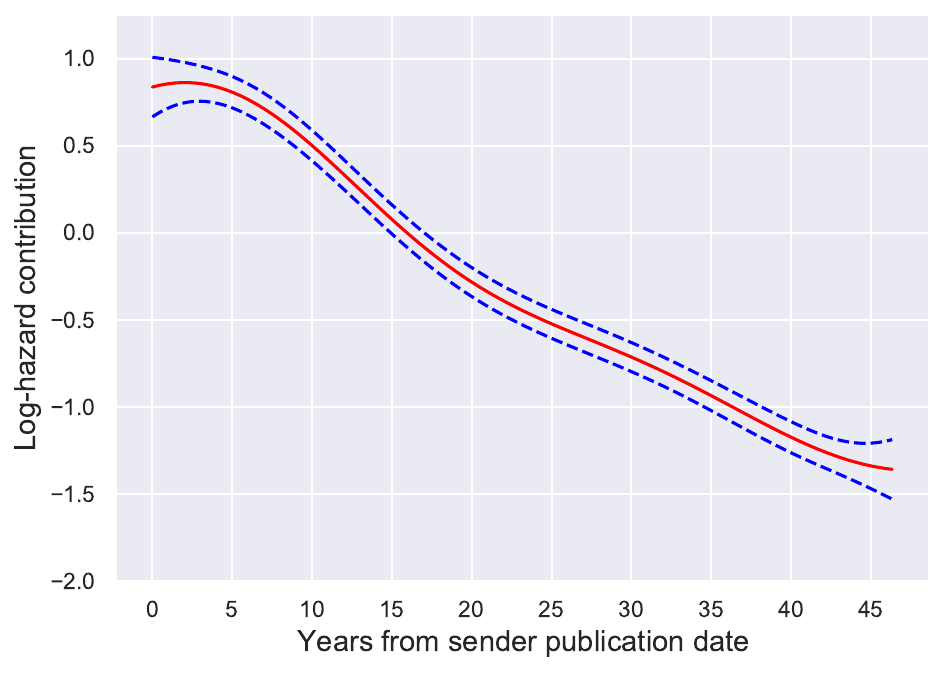}
		\caption{}
		\label{fig:lag}
	\end{subfigure}
	\begin{subfigure}{.33\textwidth}
	\centering
	\includegraphics[width=\linewidth]{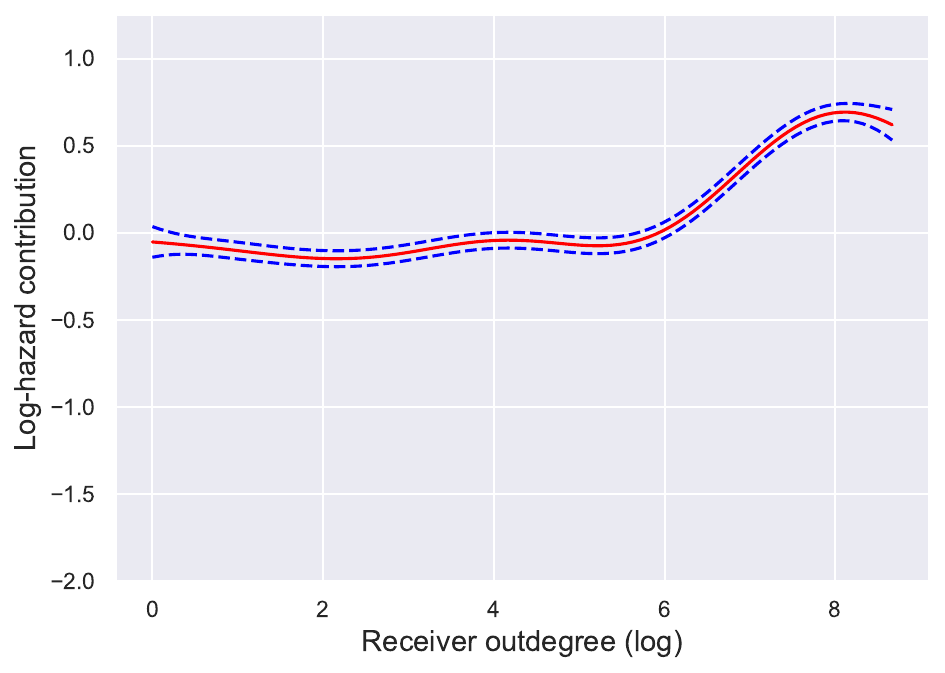}
	\caption{}
	\label{fig:outd}
\end{subfigure}
	\caption{Nodal effects consisting of a receiver publication year effect, a time-difference effect and a receiver outdegree effect.}
	\label{fig:app_eff1}
\end{figure}

\cite{filippi-mazzola_stochastic_2023} proposed to model the network via three different set of statistics: patent effects, patent similarity effects and endogenous temporal effects. The first set of effects are portrayed in Fig.~\ref{fig:app_eff1} and consists of the receiver publication year, the time-difference between the sender issue date and the receiver publication date, and the receiver outdegree. Fig.\ref{fig:pub_year} shows a maximum around the year 2000. Potentially, this indicates that increased technological innovation happend during that time. The time-difference effect identifies a period of approximately 5 years following the patent publication date when citations are most likely. Finally, the receiver outdegree confirms that patents with a higher number of citations at the time of publication tend to play a more central role in the network, consequently increasing their chances of accumulating more citations over time.


\begin{figure}
	\centering
	\begin{subfigure}{.5\textwidth}
		\centering
		\includegraphics[width=\linewidth]{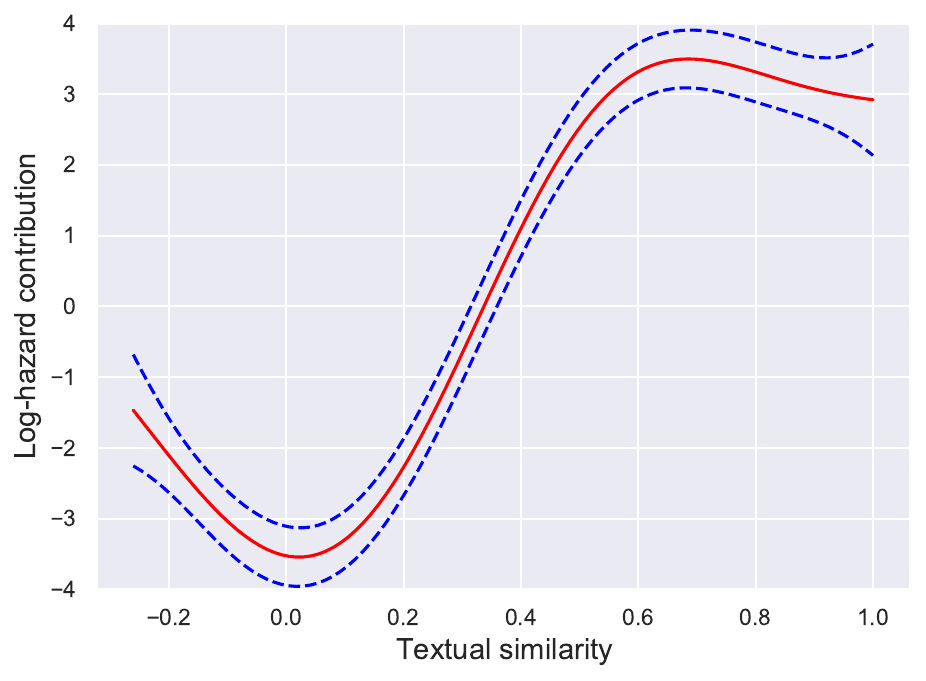}
		\caption{}
		\label{fig:sim}
	\end{subfigure}%
	\begin{subfigure}{.5\textwidth}
		\centering
		\includegraphics[width=\linewidth]{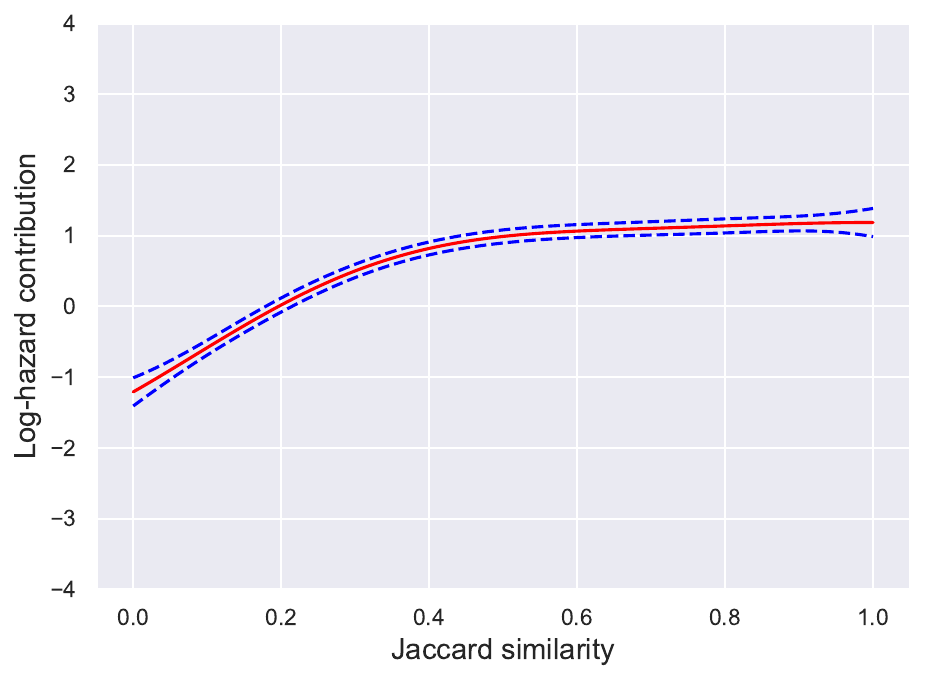}
		\caption{}
		\label{fig:jac_sim}
	\end{subfigure}
	\caption{Similarity effect consisting of the textual similarity effect and the technological relatedness effect.}
	\label{fig:app_eff2}
\end{figure}

The second set of effects, shown in Fig.~\ref{fig:app_eff2}, delves into the patent similarity characteristics that contribute to a citation. The first statistic is textual similarity. We embed patent abstracts in an euclidean space using a pre-trained SBERT model \citep{reimers-2019-sentence-bert}, and calculate pairwise cosine similarities. The resulting non-linear effects conclusively demonstrate that patents are more likely to cite each other when their abstracts share significant textual similarities. Secondly, we consider the technological relationship between the two patents, as indicated by their shared International Patent Classification (IPC) classes. We capture the proportion of shared classes to the total classes observed across both patents by computing the Jaccard similarity among these IPC classes. According to the results Fig.~\ref{fig:jac_sim}, the rate of one patent citing another increases as shared the number of technology classes increases.


\begin{figure}
	\centering
	\begin{subfigure}{.5\textwidth}
		\centering
		\includegraphics[width=\linewidth]{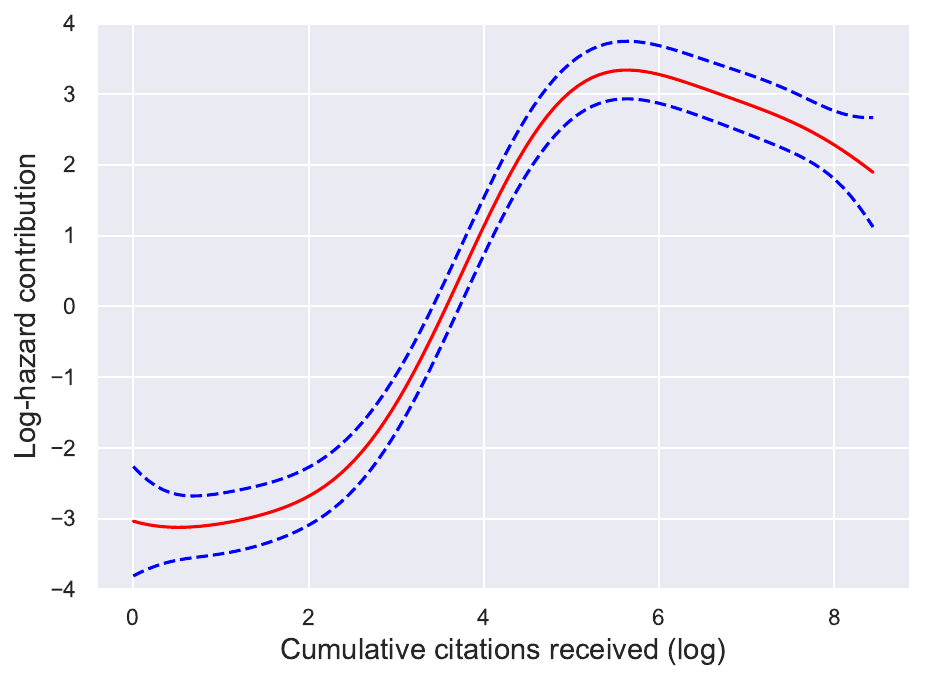}
		\caption{}
		\label{fig:cumu_cit_rec}
	\end{subfigure}%
	\begin{subfigure}{.5\textwidth}
		\centering
		\includegraphics[width=\linewidth]{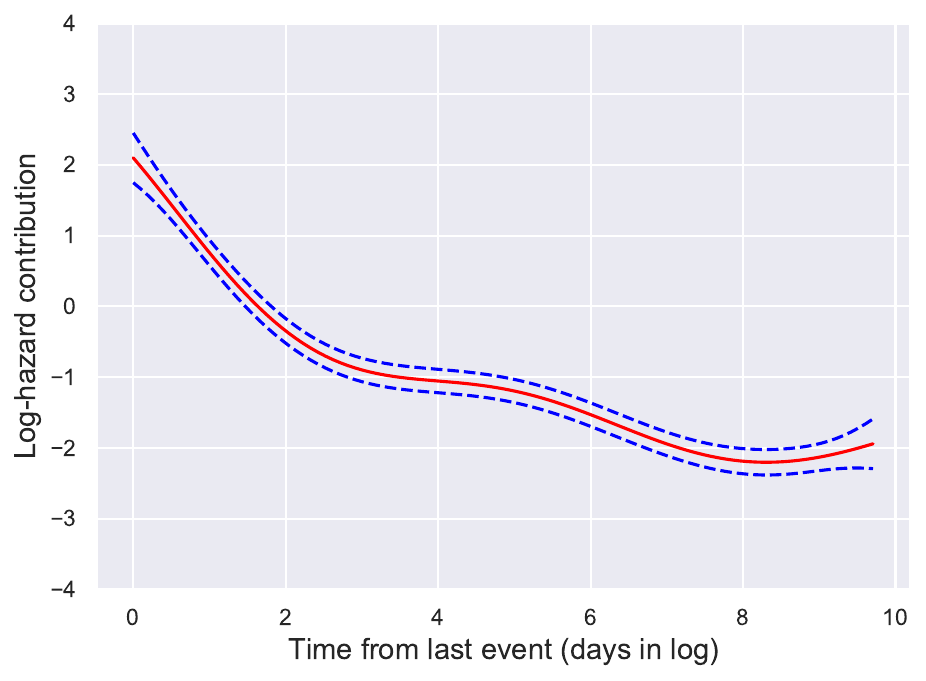}
		\caption{}
		\label{fig:tfe}
	\end{subfigure}
	\caption{Time-varying effects consisting of the cumulative citations received and the time from last event.}
	\label{fig:app_eff3}
\end{figure}

Fig.~\ref{fig:app_eff3} captures time-varying factors that influence the rate of a patent being cited. The first of these is the cumulative citations a patent has received, illustrating that as a patent accumulates more citations, its probability of receiving additional ones increases, until it reaches a plateau. The second effect evaluates the time elapsed since a patent most recent citation. This indicates that the longer the duration since the last citation, the less probable it becomes for the patent to be cited again.

\section{Conclusions}

Relational Event Models offer a versatile framework for modeling dynamic networks. Yet, their real-time application often faces challenges due to the computational complexity in their fitting procedures. Such challenges tend to amplify as the volume of observed events increases. In this study, we present a solution to these computational issues by introducing the Deep Relational Event Additive Model (DREAM). In DREAM, the non-linear behavior of each covariate is captured by an independent Artificial Neural Network, providing both precision and efficiency in capturing network dynamics. 

We proposed two distinct methods in DREAM for estimating areas of uncertainty. The first method entails non-parametrically bootstrapping the observed dataset and then refitting the model multiple times. The second method, employs Gaussian Process Regression based on a small subset of non-parametric bootstrap refits, offering a more efficient way to handle uncertainty while maintaining robustness in our estimations.


Throughout a series of simulation studies, we introduced and tested the capabilities of DREAM, emphasizing its ability in capturing non-linear effects within large dynamic networks. The robustness and efficiency of DREAM became clear when compared to existing methods such as GAMs from the MGCV package in R. DREAM strength lies not only in its ability to accurately model nonlinear effects, but also in its fast convergence, which is accomplished by leveraging the computational advantages of Pytorch and GPUs. We further demonstrated the practical significance of DREAM by modeling a patent citation network, which encompasses nearly 100 million events and about 8 million actors. 

DREAM not only offers an efficient and scalable approach to analyzing longitudinal networks and capturing complex non-linear effects, but it also offers remarkable flexibility to customize model complexity. This adaptability includes compatibility with traditional regularization methods like dropout, ridge, and lasso. Beyond its current framework, we argue that DREAM has the potential to model sophisticated non-linear interaction effects via the fitting of hyperplanes. Moreover, DREAM flexibility allows to be easily adapted to adress multi-cast interactions \citep{perry_point_2013} and further extended to model hyperedges \citep{lerner_hyper_2023}.

\section*{Acknowledgments}
This work was supported by funding from the Swiss National Science Foundation (SNSF grant 192549).

\newpage
\singlespacing			
\bibliography{reference}

\begin{thebibliography}{}

\bibitem[Agarap, 2019]{agarap_relu_2019}
Agarap, A.~F. (2019).
\newblock Deep learning using rectified linear units (relu).
\newblock abs/1803.08375.

\bibitem[Agarwal et~al., 2021]{agarwal_nam_2021}
Agarwal, R., Melnick, L., Frosst, N., Zhang, X., Lengerich, B., Caruana, R.,
  and Hinton, G. (2021).
\newblock Neural {Additive} {Models}: {Interpretable} {Machine} {Learning} with
  {Neural} {Nets}.
\newblock arXiv:2004.13912 [cs, stat].

\bibitem[Amati et~al., 2019]{amati_healthcare_2019}
Amati, V., Lomi, A., and Mascia, D. (2019).
\newblock Some days are better than others: Examining time-specific variation
  in the structuring of interorganizational relations.
\newblock {\em Social Networks}, 57:18--33.

\bibitem[Bauer et~al., 2022]{bauer_smooth_2022}
Bauer, V., Harhoff, D., and Kauermann, G. (2022).
\newblock A smooth dynamic network model for patent collaboration data.
\newblock {\em AStA Advances in Statistical Analysis}, 106(1):97--116.

\bibitem[Bianchi et~al., 2024]{bianchi_review_2023}
Bianchi, F., Filippi-Mazzola, E., Lomi, A., and Wit, E.~C. (2024).
\newblock Relational event modeling.
\newblock {\em Annual Review of Statistics and Its Application}, 11(1):null.

\bibitem[Borgan et~al., 1995]{borgan_methods_1995}
Borgan, O., Goldstein, L., and Langholz, B. (1995).
\newblock Methods for the {Analysis} of {Sampled} {Cohort} {Data} in the {Cox}
  {Proportional} {Hazards} {Model}.
\newblock {\em The Annals of Statistics}, 23(5).

\bibitem[Boschi et~al., 2023]{boschi_smooth_2023}
Boschi, M., Juozaitienė, R., and Wit, E.-J.~C. (2023).
\newblock Smooth {Alien} {Species} {Invasion} {Model} with {Random} and
  {Time}-{Varying} {Effects}.
\newblock arXiv:2304.00654 [stat].

\bibitem[Butts, 2008]{butts_2008}
Butts, C.~T. (2008).
\newblock 4. a relational event framework for social action.
\newblock {\em Sociological Methodology}, 38(1):155--200.

\bibitem[Cox, 1972]{cox_regression_1972}
Cox, D.~R. (1972).
\newblock Regression models and life-tables.
\newblock {\em Journal of the Royal Statistical Society. Series B
  (Methodological)}, 34(2):187--220.

\bibitem[Cox, 1975]{cox_partial_1975}
Cox, D.~R. (1975).
\newblock {Partial likelihood}.
\newblock {\em Biometrika}, 62(2):269--276.

\bibitem[{De Boor}, 1972]{DeBoor1972}
{De Boor}, C. (1972).
\newblock On calculating with b-splines.
\newblock {\em Journal of Approximation Theory}, 6(1):50--62.

\bibitem[Efron, 1979]{efron_boot_1979}
Efron, B. (1979).
\newblock Bootstrap methods: Another look at the jackknife.
\newblock {\em The Annals of Statistics}, 7(1):1--26.

\bibitem[Filippi-Mazzola and Wit, 2023]{filippi-mazzola_stochastic_2023}
Filippi-Mazzola, E. and Wit, E.~C. (2023).
\newblock A {Stochastic} {Gradient} {Relational} {Event} {Additive} {Model} for
  modelling {US} patent citations from 1976 until 2022.
\newblock arXiv:2303.07961 [stat].

\bibitem[Fritz et~al., 2020]{network_survey_kauerman_2020}
Fritz, C., Lebacher, M., and Kauermann, G. (2020).
\newblock Tempus volat, hora fugit: A survey of tie-oriented dynamic network
  models in discrete and continuous time.
\newblock {\em Statistica Neerlandica}, 74(3):275--299.

\bibitem[Hastie and Tibshirani, 1986]{hastie_gams_1986}
Hastie, T. and Tibshirani, R. (1986).
\newblock {Generalized Additive Models}.
\newblock {\em Statistical Science}, 1(3):297 -- 310.

\bibitem[Juozaitienė et~al., 2023]{juozaitiene_ecological_2023}
Juozaitienė, R., Seebens, H., Latombe, G., Essl, F., and Wit, E.~C. (2023).
\newblock Analysing ecological dynamics with relational event models: The case
  of biological invasions.
\newblock {\em Diversity and Distributions}, 29(10):1208--1225.

\bibitem[Kingma and Ba, 2017]{kingma_adam_2017}
Kingma, D.~P. and Ba, J. (2017).
\newblock Adam: {A} {Method} for {Stochastic} {Optimization}.
\newblock arXiv:1412.6980 [cs].

\bibitem[Kotthaus et~al., 2015]{kotthaus_runtime_2015}
Kotthaus, H., Korb, I., Lang, M., Bischl, B., Rahnenführer, J., and Marwedel,
  P. (2015).
\newblock Runtime and memory consumption analyses for machine learning {R}
  programs.
\newblock {\em Journal of Statistical Computation and Simulation},
  85(1):14--29.

\bibitem[Lerner and Lomi, 2020]{lerner_reliability_2020}
Lerner, J. and Lomi, A. (2020).
\newblock Reliability of relational event model estimates under sampling: {How}
  to fit a relational event model to 360 million dyadic events.
\newblock {\em Network Science}, 8(1):97--135.

\bibitem[Lerner and Lomi, 2023]{lerner_hyper_2023}
Lerner, J. and Lomi, A. (2023).
\newblock {Relational hyperevent models for polyadic interaction networks}.
\newblock {\em Journal of the Royal Statistical Society Series A: Statistics in
  Society}, 186(3):577--600.

\bibitem[Lomi and Bianchi, 2021]{bianchi_finance_2021}
Lomi, A. and Bianchi, F. (2021).
\newblock A time to give and a time to receive: Role switching and generalized
  exchange in a financial market.
\newblock {\em Social Networks}.

\bibitem[Lu, 2020]{Lu_dying_2020}
Lu, L. (2020).
\newblock Dying {ReLU} and initialization: Theory and numerical examples.
\newblock {\em Communications in Computational Physics}, 28(5):1671--1706.

\bibitem[Namin et~al., 2009]{Namin_tanh_2009}
Namin, A.~H., Leboeuf, K., Muscedere, R., Wu, H., and Ahmadi, M. (2009).
\newblock Efficient hardware implementation of the hyperbolic tangent sigmoid
  function.
\newblock {\em 2009 IEEE International Symposium on Circuits and Systems},
  pages 2117--2120.

\bibitem[Narayan, 1997]{Narayan_sigmoiod_1997}
Narayan, S. (1997).
\newblock The generalized sigmoid activation function: Competitive supervised
  learning.
\newblock {\em Information Sciences}, 99(1):69--82.

\bibitem[Noel et~al., 2023]{noel_gcu_2023}
Noel, M.~M., L, A., Trivedi, A., and Dutta, P. (2023).
\newblock Growing cosine unit: A novel oscillatory activation function that can
  speedup training and reduce parameters in convolutional neural networks.

\bibitem[Paszke et~al., 2019]{pytorch}
Paszke, A., Gross, S., Massa, F., Lerer, A., Bradbury, J., Chanan, G., Killeen,
  T., Lin, Z., Gimelshein, N., Antiga, L., Desmaison, A., Kopf, A., Yang, E.,
  DeVito, Z., Raison, M., Tejani, A., Chilamkurthy, S., Steiner, B., Fang, L.,
  Bai, J., and Chintala, S. (2019).
\newblock Pytorch: An imperative style, high-performance deep learning library.
\newblock In {\em Advances in Neural Information Processing Systems 32}, pages
  8024--8035. Curran Associates, Inc.

\bibitem[Patison et~al., 2015]{patison_time_2015}
Patison, K.~P., Quintane, E., Swain, D.~L., Robins, G., and Pattison, P.
  (2015).
\newblock Time is of the essence: an application of a relational event model
  for animal social networks.
\newblock {\em Behavioral Ecology and Sociobiology}, 69(5):841--855.

\bibitem[Pedregosa et~al., 2011]{scikit-learn}
Pedregosa, F., Varoquaux, G., Gramfort, A., Michel, V., Thirion, B., Grisel,
  O., Blondel, M., Prettenhofer, P., Weiss, R., Dubourg, V., Vanderplas, J.,
  Passos, A., Cournapeau, D., Brucher, M., Perrot, M., and Duchesnay, E.
  (2011).
\newblock Scikit-learn: Machine learning in {P}ython.
\newblock {\em Journal of Machine Learning Research}, 12:2825--2830.

\bibitem[Perry and Wolfe, 2013]{perry_point_2013}
Perry, P.~O. and Wolfe, P.~J. (2013).
\newblock Point process modelling for directed interaction networks.
\newblock {\em Journal of the Royal Statistical Society: Series B (Statistical
  Methodology)}, 75(5):821--849.

\bibitem[Rasmussen and Williams, 2006]{rasmussen_gaussian_2006}
Rasmussen, C.~E. and Williams, C. K.~I. (2006).
\newblock {\em Gaussian processes for machine learning}.
\newblock Adaptive computation and machine learning. MIT Press, Cambridge,
  Mass.
\newblock OCLC: ocm61285753.

\bibitem[Reddi et~al., 2018]{Reddi_adamconv_2018}
Reddi, S.~J., Kale, S., and Kumar, S. (2018).
\newblock On the convergence of adam and beyond.
\newblock {\em ArXiv}, abs/1904.09237.

\bibitem[Reimers and Gurevych, 2019]{reimers-2019-sentence-bert}
Reimers, N. and Gurevych, I. (2019).
\newblock Sentence-bert: Sentence embeddings using siamese bert-networks.
\newblock In {\em Proceedings of the 2019 Conference on Empirical Methods in
  Natural Language Processing}. Association for Computational Linguistics.

\bibitem[Ripley, 1996]{ripley_pattern_1996}
Ripley, B.~D. (1996).
\newblock {\em Pattern recognition and neural networks}.
\newblock Cambridge University Press, Cambridge ; New York.

\bibitem[Tranmer et~al., 2015]{tranmer_animal_2015}
Tranmer, M., Marcum, C.~S., Morton, F.~B., Croft, D.~P., and de~Kort, S.~R.
  (2015).
\newblock Using the relational event model ({REM}) to investigate the temporal
  dynamics of animal social networks.
\newblock {\em Animal Behaviour}, 101:99--105.

\bibitem[Vu et~al., 2017]{vu_relational_2017}
Vu, D., Lomi, A., Mascia, D., and Pallotti, F. (2017).
\newblock Relational event models for longitudinal network data with an
  application to interhospital patient transfers.
\newblock {\em Statistics in Medicine}.

\bibitem[Vu et~al., 2015]{vu_relational_2015}
Vu, D., Pattison, P., and Robins, G. (2015).
\newblock Relational event models for social learning in {MOOCs}.
\newblock {\em Social Networks}, 43:121--135.

\bibitem[Vu et~al., 2011]{vu_dynamic_2011}
Vu, D.~Q., Asuncion, A.~U., Hunter, D.~R., and Smyth, P. (2011).
\newblock Dynamic {Egocentric} {Models} for {Citation} {Networks}.
\newblock In {\em Proceedings of the 28th international conference on machine
  learning (ICML-11)}, pages 857--864.

\bibitem[Welles et~al., 2014]{welles_comunication_2014}
Welles, B.~F., Vashevko, A., Bennett, N., and Contractor, N. (2014).
\newblock Dynamic models of communication in an online friendship network.
\newblock {\em Communication Methods and Measures}, 8(4):223--243.

\bibitem[Zappa and Vu, 2021]{zappa_finance_2021}
Zappa, P. and Vu, D.~Q. (2021).
\newblock Markets as networks evolving step by step: Relational event models
  for the interbank market.
\newblock {\em Physica A: Statistical Mechanics and its Applications},
  565:125557.

\end{thebibliography}
		
\end{document}


\title{Supplementary materials for the paper ``Non parametric estimation of smooth effects in Relational Event Models with Neural Networks"}


\author[1]{Edoardo Filippi-Mazzola}
\author[1]{Ernst C. Wit}
\affil[1]{Institute of computing, Universit\`{a} della Svizzera italiana, Lugano, Swizterland}
\date{}


\renewcommand{\refname}{\large References}


\maketitle





\section*{S1 - Simulation study model selection}
\label{supp:sim_study_model_selection}
To determine the optimal configuration for DREAM,  we conducted a simulation study that examines four neural network architectures, each incorporating different degrees of dropout to apply varying levels of regularization. In this setting. the choice of dropout level is a classic trade-off between bias and variance. Too little dropout may not provide sufficient regularization, leading to overfitting. Conversely, too much dropout may lead to underfitting. Analogous to regularizing splines in GAMs, increasing penalties yield to less flexible curves, resulting in more linear representations. 

Table~\ref{tab:model_selection_sim_study} we report the four neural network architectures that have been tested. Each of these is designed with an incrementally increasing number of neurons and layers, thereby escalating the model complexity. This progressive complexity, much like the role of splines in GAMs, allows for greater flexibility in the resulting curves. The architectures range from the simplest, Model 1, with a configuration of (64, 128, 64), to the most complex, Model 4, with an expansive (512, 1024, 512, 256, 128) layout, enabling us to scrutinize the trade-off between model complexity and curve adaptability.
\begin{table}[!h]
	\centering
	\begin{tabular}{@{}ll@{}}
		\toprule
		Model & Framework \\ \midrule
		Model 1 & (64, 128, 64) \\
		Model 2 & (128, 256, 64) \\
		Model 3 & (256, 512, 256, 128) \\
		Model 4 & (512, 1024, 512, 256, 128) \\ \bottomrule
	\end{tabular}
	\caption{Summary of neural network frameworks with varying complexities.}
	\label{tab:model_selection_sim_study}
\end{table}

\begin{figure}[!h]
	\centering
	\includegraphics[width=\textwidth]{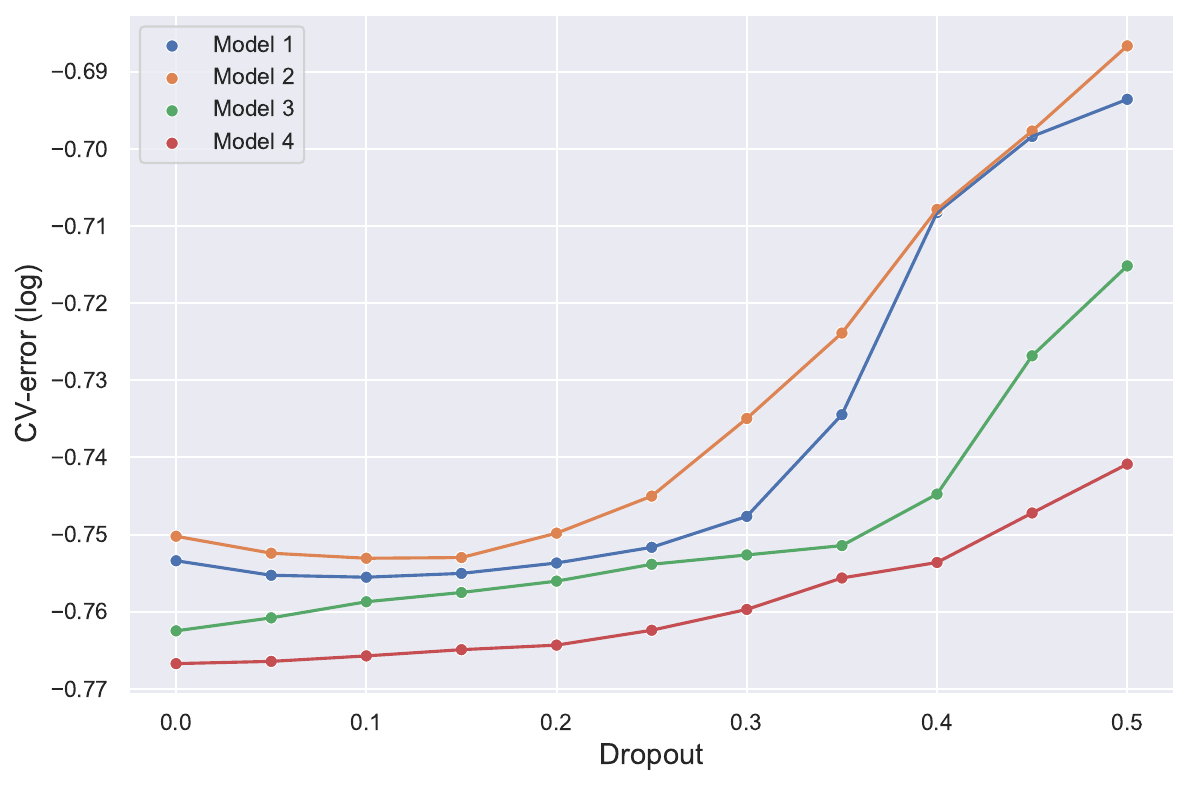}
	\caption{Results from the simulation study illustrating the performance of various model configurations.}
	\label{fig:model_selection_sim_study}
\end{figure}

Examining the insights provided by Figure~\ref{fig:model_selection_sim_study}, a clear pattern emerges indicating that the model with the highest complexity is most adequate in capturing the non-linerities. While a comparison within Model 4 reveals minimal variance between dropout levels of 0 and 0.05, our preference inclines towards the model iteration with no penalization. By avoiding the regularization imposed by dropout, we aim to preserve the model sensitivity in capturing more data, ensuring a more accurate interpretation of the underlying dynamics.

\section*{S2 - US patent citation network model selection}
\label{supp:patents_model_selection}

\begin{figure}[!h]
	\centering
	\includegraphics[width=\textwidth]{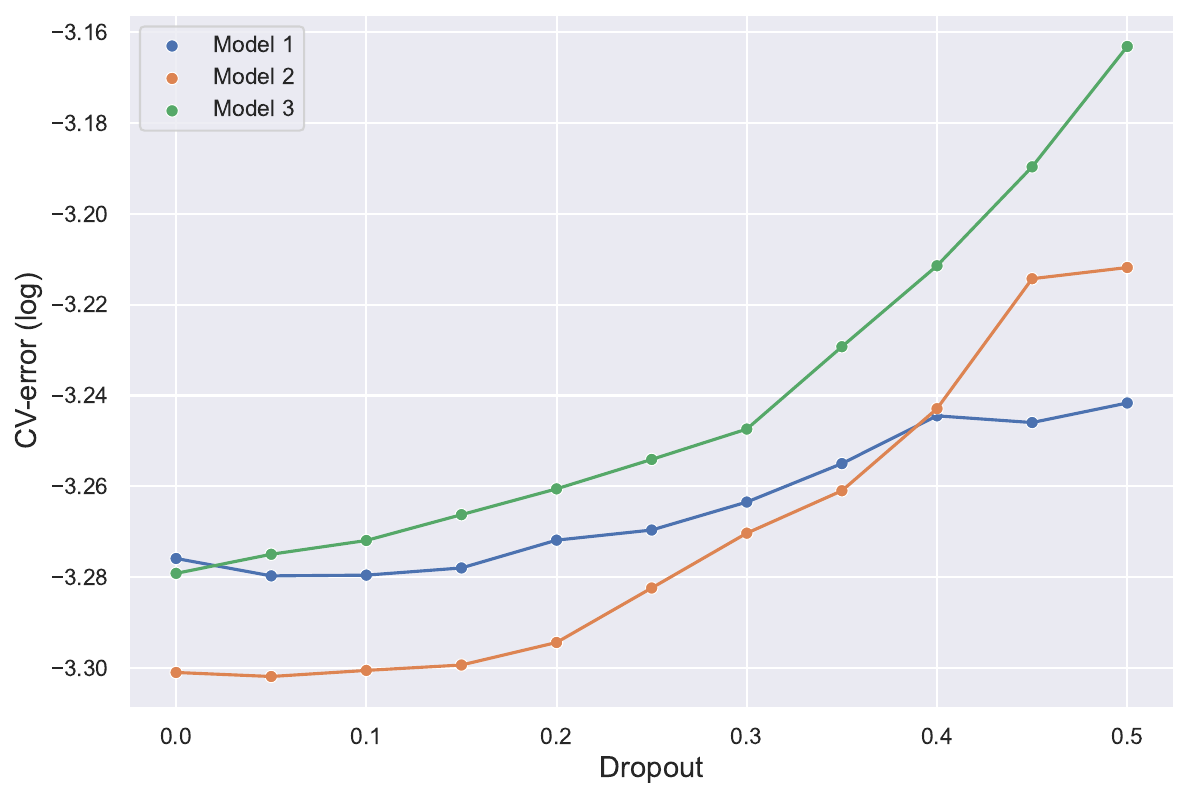}
	\caption{Cross-validation Performance Spread for Model Selection in the U.S. Patent Citation Network.}
	\label{fig:model_selection_patents}
\end{figure}

Building upon our prior simulation study, detailed in Section~\ref{supp:sim_study_model_selection}, we extend our analysis to the U.S. Patent citation network. Utilizing the same architectural frameworks, we conducted a simulation study to assess performance across the network. The comparative results are visually represented in Figure~\ref{fig:model_selection_patents}, where the spread of performance metrics is evaluated using a 10-fold cross-validation method. 

From our analysis, it becomes evident that Model 2 outperforms the others. Notably, this model demonstrates how increased computational complexity correlates with elevated cross-validation errors. Given this insight, we decided not to proceed with Model 4. The results indicated diminishing returns with higher complexity, thus reinforcing our decision to halt further simulations at Model 3. Consequently, Model 2 with a dropout of 0.05 was selected for our application due to its optimal balance of complexity and error minimization.

\newpage